\documentclass{article}

\usepackage{arxiv}

\usepackage[T1]{fontenc}    
\usepackage{hyperref}       
\usepackage{booktabs}       
\usepackage{amsfonts}       
\usepackage{nicefrac}       
\usepackage{microtype}      
\usepackage{scalefnt} 
\usepackage{multirow} 
\usepackage{graphicx}
\usepackage[square,numbers]{natbib}
\usepackage{doi}
\usepackage{setspace}

\title{Exploring traditional machine learning for identification of pathological auscultations}

\date{August 31, 2022}

\author{
\href{https://orcid.org/0000-0002-5391-4004}{\includegraphics[scale=0.08]{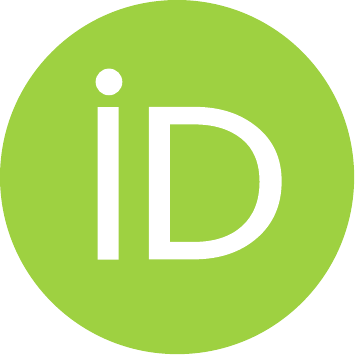}\hspace{1mm}Haroldas Razvadauskas} \\
Department of Internal Diseases \\
Faculty of Medicine \\
Lithuanian University of Health Sciences \\
Kaunas, Lithuania \\
\And
\href{https://orcid.org/0000-0002-4769-4527}{\includegraphics[scale=0.08]{orcid.pdf}\hspace{1mm}Evaldas Vaičiukynas}\thanks{Corresponding author (\texttt{evaldas.vaiciukynas@ktu.lt}), preprint submitted to Biomedical Signal Processing and Control }\\
Department of Information Systems \\
Faculty of Informatics \\
Kaunas University of Technology \\
Kaunas, Lithuania \\
\And
\href{https://orcid.org/0000-0003-0710-6791}{\includegraphics[scale=0.08]{orcid.pdf}\hspace{1mm}Kazimieras Buškus} \\
Faculty of Mathematics and Natural Sciences \\
Kaunas University of Technology \\
Kaunas, Lithuania \\
\And
Lukas Drukteinis \\
Faculty of Mathematics and Natural Sciences \\
Kaunas University of Technology \\
Kaunas, Lithuania \\
\AND
Lukas Arlauskas \\
Department of Information Systems \\
Faculty of Informatics \\
Kaunas University of Technology \\
Kaunas, Lithuania \\
\And
\href{https://orcid.org/0000-0001-6705-6154}{\includegraphics[scale=0.08]{orcid.pdf}\hspace{1mm}Saulius Sadauskas} \\
Department of Internal Diseases \\
Faculty of Medicine \\
Lithuanian University of Health Sciences \\
Kaunas, Lithuania \\
\And
\href{https://orcid.org/0000-0003-0155-6628}{\includegraphics[scale=0.08]{orcid.pdf}\hspace{1mm}Albinas Naudžiūnas} \\
Department of Internal Diseases \\
Faculty of Medicine \\
Lithuanian University of Health Sciences \\
Kaunas, Lithuania \\
}

\renewcommand{\shorttitle}{Traditional machine learning for identification of pathological auscultations}

\hypersetup{
pdftitle={Exploring traditional machine learning for identification of pathological auscultations},
pdfsubject={arXiv preprint},
pdfauthor={Haroldas Razvadauskas, Evaldas Vaiciukynas, Kazimieras Buskus, Lukas Drukteinis, Lukas Arlauskas, Saulius Sadauskas, Albinas Naudziunas},
pdfkeywords={digital health, auscultation, lung sounds, random forest, isolation forest, Surfboard},
}

\begin{document}
\maketitle

\begin{abstract}
Today, data collection has improved in various areas, and the medical domain is no exception. Auscultation, as an important diagnostic technique for physicians, due to the progress and availability of digital stethoscopes, lends itself well to applications of machine learning. Due to the large number of auscultations performed, the availability of data opens up an opportunity for more effective analysis of sounds where prognostic accuracy even among experts remains low. In this study, digital 6-channel auscultations of 45 patients were used in various machine learning scenarios, with the aim of distinguishing between normal and anomalous pulmonary sounds. Audio features (such as fundamental frequencies F0-4, loudness, HNR, DFA, as well as descriptive statistics of log energy, RMS and MFCC) were extracted using the Python library \emph{Surfboard}. Windowing and feature aggregation and concatenation strategies were used to prepare data for tree-based ensemble models in unsupervised (fair-cut forest) and supervised (random forest) machine learning settings. The evaluation was carried out using 9-fold stratified cross-validation repeated 30 times. Decision fusion by averaging outputs for a subject was tested and found to be useful. Supervised models showed a consistent advantage over unsupervised ones, achieving mean AUC ROC of 0.691 (accuracy 71.11\%, Kappa 0.416, F1-score 0.771) in side-based detection and mean AUC ROC of 0.721 (accuracy 68.89\%, Kappa 0.371, F1-score 0.650) in patient-based detection.
\end{abstract}


\onehalfspacing

\section{Introduction}

Auscultation is one of the four pillars of patient examination \citep{Narula_2018,Pereira_2017,Garvick_2022}, and is particularly important during cardiovascular and lung checks. Today, the assessment of auscultation sounds remains subjective and depends on the qualifications and expertise of a specialist\citep{Hafke_Dys_2019,Kim_2021}. \citet{Arts_2020} recommends leaving the practice of traditional auscultation to experienced specialists, since sensitivity is as low as 37\% when identifying between normal cases and several lung pathologies. Furthermore, auscultation skills require continuous practice and can even degrade over time \citep{Vukanovic_Criley_2006}, but can be improved through training \citep{Bernardi_2019}. The invention of a digital stethoscope has helped improve noise resistance and amplify lung and heart sounds \citep{Rennoll_2021}, resulting in superior sound quality, at least for 65\% patients \citep{Silverman_2019}. Furthermore, \citet{McLane_2021} propose additional techniques to remove ambient noise in lung auscultations. There exists some evidence that digital stethoscopes have the potential to increase the accuracy of evaluation if they are more widely accepted in clinical practice \citep{Kalinauskien__2019}. Remote auscultations in tele-medicine \citep{Lv_2021} is another useful application of digital stethoscopes, which was recently recommended for the monitoring of quarantined patients with COVID-19 \citep{Haskel_2022}. Although the adoption of the digital stethoscope has remained rather slow, for a variety of reasons, the evaluation of auscultation sounds still greatly depends on the skill and experience of the physician.

Therefore, the combination of electronic stethoscopes with artificial intelligence has the potential to update a 200-year-old technique \citep{Silverman_2019} by increasing detection accuracy, as well as sensitivity and specificity. It can also help mitigate differences in diagnoses between physicians and improve the speed of patient examinations \citep{Pereira_2017,Bernardi_2019,Huang_2020,Zhu_2022}.

In this research, a set of 370 acoustic indicators was estimated from the auscultation recordings, followed by supervised (pathology detection) and unsupervised (anomaly detection) machine learning variants using tree-based ensemble methods. For the experiments, the audio signals were windowed, and we have implemented feature aggregation and concatenation strategies, as well as simple decision fusion. The success of machine learning was evaluated using repeated stratified cross-validation, and various detection accuracy metrics were reported to help choose between the proposed approaches for the detection of pathological auscultations.

\section{Related work}

Due to the issues mentioned above and the large variation in the accuracy of the assessment of auscultation by specialists \citep{Grzywalski_2019a}, new approaches are being explored with the aim of reducing or eliminating the subjectivity of practice. A nice overview of research in the field of cardiac auscultation is provided by \citet{Jani_2021} and \citep{Brites_2022}. Machine learning techniques of various novelty are increasingly applied for the automated screening of heart \citep{Rubin_2017,Latif_2018,Yadav_2019,Chorba_2021,Shuvo_2021a,Lv_2021,Jeong_2021} or lung \citep{Grzywalski_2019a,Hsu_2021,Zhang_2021,Horimasu_2021,Kim_2021,Pham_2021,Shuvo_2021,Fernando_2022,Sankararaman_2022} pathologies from auscultation recordings.

\citet{Grzywalski_2019a} train a convolutional neural network model (StethoMe AI) using spectrograms to classify waveform segments into four categories (wheezes, rhonchi, fine, and coarse crackles). When comparing several accuracy metrics (sensitivity / recall, specificity, precision, and F1-score) between the model and a group of five physicians, \citet{Grzywalski_2019a} reports consistently better performance for machine learning model with a mean F1-score of 0.63 versus 0.54 for physicians. Other researchers also devise various deep learning solutions, demonstrating mean F1-scores as high as 0.85 \citep{Pham_2021} or 0.92 \citep{Fernando_2022} or even 0.99 \citep{Shuvo_2021} on the \emph{ICBHI 2017} scientific challenge respiratory sound database\cite{Rocha_2017}. \citet{Kim_2021} collected their own dataset and used convolutional neural network achieving an AUC ROC of 0.93 when discriminating between normal and abnormal breath sounds (crackles, wheezes, and rhonchi). \citet{Hsu_2021} recently introduced the largest open access dataset \emph{HF\_Lung\_V1} and using combination of recurrent and convolutional neural architectures reported detection success by AUC ROC of 0.86 - 0.96, which was soon outperformed by \citet{Fernando_2022} using lightweight multi-branch temporal convolution networks. \citet{Sankararaman_2022} form his own dataset of 30 bronchial and 30 stridor breath sounds from several different sources and, potentially due to the sounds being clear examples of the respective class and using test data in graph construction, demonstrate that graph-based features, derived from the correlation mapping approach, projected into 2D with principal component analysis help to perfectly separate classes with the linear discriminant analysis model achieving 100\% accuracy. \citet{Zhang_2021} use a proprietary breath analysis algorithm to discriminate between crackles, wheezes and normal breath and report a mean F1-score of 0.81 versus 0.58 for pediatricians. \citet{Horimasu_2021} achieved AUC ROC of 0.86 for 60 patients using the fine crackles quantitative value (FCQV) marker, which is calculated using a machine-learned polynomial consisting of a hundred or more features obtained by frequency, local variance, and cepstrum analyzes.

\section{Auscultation recordings}

Subjects were recorded with the Kaunas Regional Bioethics Commitee permit (2021-05-11 No. BE-2-57). Auscultations were recorded in a clinical setting using a 3M Littmann 3200 stethoscope (IEC60601-1-2). The recordings were made in the wards of the Departments of Cardiology and Internal Medicine of Kaunas Hospital at the Lithuanian University of Health Sciences. Therefore, recordings were not free of additional background noise and non-pulmonary sounds, but noise canceling mode was used on the stethoscope.

During auscultation, the patient was breathing through his or her mouth, and each recording covered approximately 3 to 4 breathing cycles, where a single breathing cycle corresponds to complete inspiration and expiration of air by the patient. Slightly more healthy subjects and more males were auscultated, as can be seen in Table \ref{tab:dataset}.

\begin{table}[!htb]
\scalefont{0.85}
\centering
\caption{Pivot table of auscultated subjects by class and sex with age mean and standard deviation (in parentheses).}
\label{tab:dataset}
\begin{tabular}{@{}lcccccc@{}}
\toprule
\bf{Pathology} & \bf{Female count} & \bf{Female age} & \bf{Male count} & \bf{Male age} & \bf{Overall count} & \bf{Overall age} \\ \midrule
No      & 12 & 66.00 (20.32) & 14 & 49.43 (19.45) & 26 & 57.08 (21.50) \\
Yes     & 8 & 71.88 (11.05) & 11 & 73.82 (11.18) & 19 & 73.00 (11.17) \\
\bf{Overall} & 20 & 68.35 (17.46) & 25 & 60.16 (20.33) & 45 & 63.80 (19.54) \\ \bottomrule
\end{tabular}
\end{table}

\begin{figure}[!htb]
\centering
\includegraphics[width=.5\linewidth]{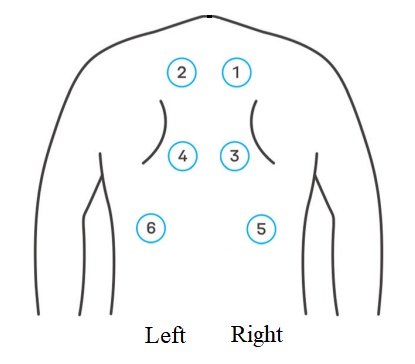}
\caption{\label{fig:localizations} The specific locations of the auscultation points on the individual subject's back, corresponding to 6 channels. Illustration adapted from \citet{Grzywalski_2019a}.}
\end{figure}

The length of the recording in .wav format (mono PCM, 16 bits at 4 kHz sampling rate) was exactly 15 seconds. The data collected consisted of six audio recordings for each subject, corresponding to six channels (physical body locations - points); therefore, 45 patients resulted in a sound corpora of 270 recordings. The six recorded points were from the back of the subject's thorax (see Fig. \ref{fig:localizations}): two points were located above the scapula (right and left), two between the scapula (right and left) and two below the scapula, where the periphery of the lungs was auscultated (right and left).

\section{Methodology}

In this study, we used six-channel audio data collected by a digital auscultation device to explore the effectiveness of the traditional machine learning algorithm to classify normal and abnormal lung sounds. Random forests were used in a supervised setting, and isolation forests were used in an unsupervised setting.

\subsection{Data preparation}

Three variants of the audio corpus were prepared: original recordings of 15 s duration (\emph{w0}), recordings splitted into 3 (\emph{w3}) and 5 (\emph{w5}) parts. The splitting was carried out using a sliding window with 50\% overlap; therefore, the resulting .wav files in the \emph{w3} dataset were of 7.5 s duration and .wav files in \emph{w5} dataset were of 5 s duration. Windowing provides an opportunity to make additional data transformations and aggregate features (using mean and standard deviation), overlapping ensures that some audio information is not lost in windowing, and differing number of windows enables broader experimentation.

For each .wav file in non-windowed (\emph{w0}) or windowed (\emph{w3} and \emph{w5}) audio corpuses, 370 audio features were calculated, constituting a feature vector for machine learning. Features were calculated using the audio feature extractor library \emph{Surfboard} \citep{lenain2020surfboard} in Python. The collection was based on 377 audio features recommended by the authors and related to the domain of Parkison detection from voice / speech (in \emph{parkinsons\_features.yaml} configuration file), with some features (F0 contour, jitter, shimmer and PPE) having to be excluded due to the excessive amounts of \emph{NA (not available)} values. The remaining set of audio features consisted of fundamental frequencies F0-4, loudness, HNR, DFA, as well as descriptive statistics of log energy, RMS and MFCC.

For experiments, three corpuses were pre-processed as follows: 
\begin{itemize}
	\item raw features (\emph{raw}) for channel-level (\emph{w0}) or window-level (\emph{w3} and \emph{w5}) data, resulting in datasets having 370 columns (audio features) and differing number of rows (270, 810 or 1350 cases); 
	\item aggregated features (\emph{cms} and \emph{wms}) by obtaining mean and standard deviation of each audio feature for all channels (\emph{cms}) or all windows of a single channel (\emph{wms} for \emph{w3} and \emph{w5} corpuses), which increases feature vector from 370 to 740 elements;
	\item concatenated features (\emph{c2}, \emph{c3}, and \emph{c6}) by stacking feature vectors into a long vector with information from several channels available to the model at once to induce channel-specific internal model representations when feature vector is expanded by joining side by side either left-right (\emph{c2}), upper-middle-lower (\emph{c3}) or all (\emph{c6}) channels.
\end{itemize}

For unsupervised machine learning, the number of features in the datasets ranged from 370 to 4440 without and from 370 to 2220 with decision fusion, see Table \ref{tab:prepared}. For supervised machine learning, datasets were additionally enriched with meta-information on channel (1-6), side (right or left), and level (upper, middle or lower), where applicable. Age and sex were not used as features in the experiments, although sex categories were used together with diagnosis for stratification purposes in cross-validation to obtain a more fair split of subjects into train and test sets.

\begin{table}[!htb]
\scalefont{0.85}
\centering
\caption{Datasets prepared for machine learning with various detection scopes. The size of the dataset, in terms of the number of rows and columns for an unsupervised model, is reported in parentheses. The number of rows for the same level scope without decision fusion is constant, whereas with decision fusion the number of rows increases several times (at least twice).}
\label{tab:prepared}
\begin{tabular}{@{}lp{0.4\textwidth}p{0.4\textwidth}@{}}
\toprule
\bf{Detection scope} & \bf{Without decision fusion}                     & \bf{With decision fusion} \\ \midrule
patient-based &
  w0 cms ( 45 x 740 ), w3 cms ( 45 x 740 ), w5 cms ( 45 x 740 ), w0 c6 ( 45   x 2220 ), w5 c6 ( 45 x 4440 ) &
  w0 raw ( 270 x 370 ), w3 raw ( 810 x 370 ), w5 raw ( 1350 x 370 ), w5 wms ( 270 x 740 ), w0 c2 ( 135 x 740 ), w0 c3 ( 90 x 1110 ), w5 c2 ( 135 x 1480   ), w5 c3 ( 90 x 2220 ) \\
side-based      & w0 c3 ( 90 x 1110 ), w5 c3 ( 90 x 2220 )    & w0 raw ( 270 x 370 ), w3 raw ( 810 x 370 ), w5 raw ( 1350 x 370 ), w5 wms   ( 270 x 740 ) \\
level-based     & w0 c2 ( 135 x 740 ), w5 c2 ( 135 x 1480 )   & w0 raw ( 270 x 370 ), w3 raw ( 810 x 370 ), w5 raw ( 1350 x 370 ), w5 wms   ( 270 x 740 ) \\
channel-based   & w0 raw ( 270 x 370 ), w5 wms ( 270 x 740 )  & w3 raw ( 810 x 370 ), w5 raw ( 1350 x 370 )                                               \\
window-based    & w3 raw ( 810 x 370 ), w5 raw ( 1350 x 370 ) &  \\ \bottomrule
\end{tabular}
\end{table}

\paragraph{Decision fusion.} Decision-level fusion, in general, corresponds to an ensemble of classifiers: the individual outputs of each model are combined in a meta-learner fashion to achieve a better final prediction. Different model types using the same feature set or the same type of model using distinct feature sets are commonly used to achieve diversity. In our work, meta-learner was not used; instead, model outputs (predictions), belonging to the same subject, were simply averaged to arrive at the final decision. The averaging was carried out by and the decisions were fused for the patient code to obtain patient-based detection, the side (left and right) to obtain side-based detection, the level (upper, middle, and lower) to obtain level-based detection, the channel (from 1 to 6) to obtain channel-based detection, and the window (either 3 or 5) to obtain window-based detection.

\subsection{Random forest}

The random forest \citep{Breiman_2001} (RF) is a supervised machine learning algorithm, an efficient and effective classifier in many use cases, based on a CART-type decision tree ensemble. RF classifier decides by pooling a committee of many decision trees, built on different bootstrapped samples of a dataset and a randomized subset of features in it. Such an approach provides RF qualities such as robustness against over-fitting, efficiency, and generalization error converging as the number of trees increases. To achieve low bias and low correlation among individual trees, which are essential for the ensemble to accurately classify data, trees are grown to maximum depth (bias) and randomized (correlation) as follows:

\begin{itemize}
\item Each tree of RF is grown on the data obtained by bootstrapping sample on the given dataset;
\item In the growing process, for each node, the \emph{mtry} variables are selected at random and, with respect to the best split metric, a single variable is chosen for the node.
\end{itemize}

In our study, the random forest (R package \emph{ranger}) was set to always split by meta-information variables, such as side (left-right position), level (upper-middle-lower position) and channel (exact location from 1 to 6), depending on the dataset used.

\subsection{Isolation forest}

The isolation forest \citep{Liu_2008} is an unsupervised machine learning algorithm, considered as a further development of the decision tree ensemble method, built to isolate and detect anomalies in the given data. Since profiling ‘normality’ in the data would require linear time and memory complexity in most cases, the isolation forest takes on the analysis of outliers, relying on the fact that the anomalies in the data are few and they look different. More concretely, the algorithm recursively partitions given data by selecting random features and splitting values in the range of feature observations in the dataset, thus creating the so-called isolation trees (iTree). Because the splits made for observations to be isolated could be different for each sample, this difference in path lengths will signal the outliers, with more extreme values being closer to the root node, thus having a shorter path length. In an isolation forest, the average depth is calculated for all trees in the ensemble and anomalies are classified as samples that reside higher in the trees and have a lower average depth. In our work, the implementation of the isolation tree in the R library \emph{isotree} \citep{isotreeLib} with the parameter \emph{prob\_pick\_pooled\_gain} set to 1, indicating the probability of choosing the threshold to divide a linear combination of several variables as the threshold that maximizes the criterion of pooled standard deviation gain \citep{cortes2019distance,imparxiv} of the combination. These settings result in the fair-cut forest (FCF) variant \citep{faircut} of the isolation forest model.

\subsection{Model evaluation}

Machine learning success was evaluated using k-fold stratified cross-validation (CV) repeated 30 times, where subjects are split into approximately equal-sized groups, and the model is trained on all parts except one that is left out to test model inference and calculate accuracy. To extensively evaluate detection performance, the following metrics were used, such as

\begin{itemize}
    \item Area under the receiver operating characteristic curve (AUC ROC), which corresponds to the probability that a randomly chosen negative instance will have a smaller estimated probability of belonging to the positive class than a randomly chosen positive instance \citep{Jin_Huang_2005}. In short, AUC summarizes probability of correctly ranking a (normal, pathological) pair of data examples based on a detector output and is directly related to Wilcoxon Mann-Whitney statistic
    \item Precision-Recall curve also allows calculating area under the curve (AUC ROC) and the larger class imbalance the more useful Precision-Recall curve can be over ROC.
    \item Accuracy (Acc) - proportion of overall correct classifications:
     \begin{equation}
	    \frac {TP + TN}{TP + FP + TN + FN}
    \end{equation}
   \item  Sensitivity (Sens, Recall) - target class recall:
   \begin{equation}
	\frac {TP}{TP + FN}
    \end{equation}
   \item  Specificity (Spec) - non-target class recall:
   \begin{equation}
	\frac {TN}{TN + FP}
    \end{equation}
    \item Precision (Prec) - target class precision:
   \begin{equation}
	\frac {TP}{TP + FP}
    \end{equation}
    \item Negative predictive value (NPV) - non-target class precision:
   \begin{equation}
	\frac {TN}{TN + FN}	
    \end{equation}
     \item F1-score (F1) - harmonic mean of target class precision and recall:
   \begin{equation}
	2 \times \frac {Precision \times Recall} {Precision + Recall}
    \end{equation}
     \item  Kappa \citep{cohkappa,McHugh_2012} - accuracy, corrected for class imbalance:
   \begin{equation}
	    \frac {p_0 - p_e} {1 - p_e}
	\end{equation}
	where \emph{p\textsubscript{e}} – sum of probabilities of predictions agreeing with the ground-truth by chance, \emph{p\textsubscript{0}} – overall accuracy of the model.
\end{itemize}

In the formulas above, TP, TN, FP, and FN denote true positive, true negative, false positive, and false negative, respectively, when using absolute frequencies (counts) from the confusion matrix, obtained by thresholding model's output prediction (naive value being 0.5, but another more effective operating point could be selected).

Plot-based AUC ROC and AUC PRC metrics are calculated from raw outputs of the models, whereas to calculate the remaining metrics one needs to obtain confusion matrix by using threshold on raw outputs for converting soft decision to hard decision. The optimal threshold was estimated at an operating point that gave an equal error rate instead of using the default value of 0.5.

\section{Experiments}

Various approaches to build a detector for pathological auscultation recordings were evaluated by repeating the 9-fold cross-validation 30 times. Cross-validation (\emph{createMultipleFolds} function in the R package \emph{caret}) was stratified using a combination of subject's sex and diagnosis (class label), ensuring the same proportions of sex / diagnosis in both training and testing folds. Models were built using training fold data and predictions on testing fold data were saved. After inference on all 9 testing folds, the obtained predictions were pooled for comparison with the ground-truth, enabling calculation of various evaluation metrics. The mean of AUC ROC and AUC PRC, estimated by repeating the 9-fold cross-validation 30 times, together with its 90\% confidence interval are reported in the results tables.

Additional evaluation metrics, calculated from the confusion matrix, were obtained by finding the central run (the closest to mean AUC ROC) of all 30 runs and choosing the threshold at the equal error rate operating point (\emph{Sens=Spec} heuristic in R package \emph{PresenceAbsence}), which is the point where a diagonal crosses the ROC curve. The useful overall accuracy (Acc) should exceed the null accuracy of 57.77\% (proportion of the majority class), and the Kappa metric helps to account for that, with its values in the range 0.2 - 0.4 already considered a fair agreement \citep{McHugh_2012}. Figures compare the best results between unsupervised and supervised machine learning settings using mean ROC and PRC curves with 90\% confidence interval in gray (using the R package \emph{precrec}).

In unsupervised machine learning with isolation forest (R package \emph{isotree}), the model was created on all data of the training fold using fixed hyperparameters (\emph{ntrees=500, ndim=3, prob\_pick\_pooled\_gain=1}), resulting in a fair-cut forest model (FCF).

In supervised machine learning with random forest (RF), only the number of trees was fixed (\emph{num.trees=500}) and training fold data were used to tune the main hyperparameters (\emph{min.node.size} and \emph{mtry}) of the model by minimizing the log-loss metric of the bag out of the bag (R packages \emph{ranger}, \emph{mlr3} and \emph{tuneRanger}). The total number of tuning iterations was 30 with 19 for warm-up (trying random values of hyper-parameters) and 11 for tuning (through Bayesian optimization). The rather useful parameter of \emph{ranger} random forest is \emph{always.split.variables}, which enforces considering some features to be included for every node of the tree and was set to meta-information features (side, channel, and level) with several exceptions: for \emph{RF w0 c6} and \emph{RF w5 c6} it was set to \emph{NULL}, for \emph{RF w0 c3} and \emph{RF w5 c3} it was set to \emph{Side}, and for \emph{RF w0 c2} and \emph{RF w5 c2} it was set to \emph{Level}.

\subsection{Patient-based detection results}

Patient-based results constitute detection accuracy when a single final decision is obtained for a subject. Accuracy here is understood as a percentage of correctly predicted subjects when the task is to distinguish between healthy and pathological cases.

In the unsupervised setting (see Table~\ref{tab:unsup-patient}), the best overall result was a mean AUC ROC of 0.622 without and a mean AUC ROC of 0.568 with decision fusion. The winning solution without fusion was \emph{FCF w5 cms}, which uses mean and standard deviation statistics of 30 values (all 6 channels and 5 windows of each channel) for each audio feature, resulting in 740 features (370 means and 370 standard deviations). The winning solution with fusion was \emph{FCF w0 c3}, which uses feature-level fusion by simple concatenation of features from 3 channels side by side and results in separate predictions for the right and left sides, then both predictions are averaged to form a final prediction for the subject. Interestingly, half of the decision fusion solutions in an unsupervised setting - \emph{FCF w0 raw}, \emph{FCF w3 raw}, \emph{FCF w0 c2} – were rather close to the winner \emph{FCF w0 c3}.

\begin{table}[!htb]
\scalefont{0.85}
\caption{\label{tab:unsup-patient}Patient-based results in the unsupervised machine learning setting. Notes: bold denotes the maximum value of the column, and decision fusion term 'fused' corresponds to 'fused for Code'.}
\centering
\begin{tabular}[t]{lccccccccc}
\toprule
\bf{Model} & \bf{AUC ROC} & \bf{AUC PRC} & \bf{Acc} & \bf{Kappa} & \bf{Sens} & \bf{Spec} & \bf{Prec} & \bf{NPV} & \bf{F1} \\
\midrule
FCF w0 cms ( 45 x 740 ) & 0.540 ± 0.005 & 0.559 ± 0.003 & 51.11 & 0.026 & 0.526 & \bf{0.500} & 0.435 & 0.591 & 0.476\\
FCF w3 cms ( 45 x 740 ) & 0.581 ± 0.007 & 0.582 ± 0.005 & 48.89 & 0.015 & 0.632 & 0.385 & 0.429 & 0.588 & 0.511\\
FCF w5 cms ( 45 x 740 ) & \bf{0.622 ± 0.006} & \bf{0.612 ± 0.005} & \bf{55.56} & \bf{0.149} & \bf{0.737} & 0.423 & \bf{0.483} & \bf{0.688} & \bf{0.583}\\
FCF w0 c6 ( 45 x 2220 ) & 0.557 ± 0.007 & 0.556 ± 0.007 & 46.67 & -0.049 & 0.526 & 0.423 & 0.400 & 0.550 & 0.455\\
FCF w5 c6 ( 45 x 4440 ) & 0.498 ± 0.009 & 0.504 ± 0.010 & 44.44 & -0.085 & 0.526 & 0.385 & 0.385 & 0.526 & 0.444\\
\midrule
FCF w0 raw ( 270 x 370 ) fused & 0.539 ± 0.005 & 0.591 ± 0.005 & 46.67 & -0.034 & 0.579 & 0.385 & 0.407 & 0.556 & 0.478\\
FCF w3 raw ( 810 x 370 ) fused & 0.553 ± 0.005 & 0.604 ± 0.005 & 51.11 & 0.012 & 0.474 & 0.538 & 0.429 & 0.583 & 0.450\\
FCF w5 raw ( 1350 x 370 ) fused & 0.563 ± 0.005 & \bf{0.606 ± 0.003} & 46.67 & -0.007 & \bf{0.684} & 0.308 & 0.419 & 0.571 & \bf{0.520}\\
FCF w5 wms ( 270 x 740 ) fused & 0.496 ± 0.003 & 0.521 ± 0.003 & 51.11 & 0.012 & 0.474 & 0.538 & 0.429 & 0.583 & 0.450\\
FCF w0 c2 ( 135 x 740 ) fused & 0.528 ± 0.006 & 0.559 ± 0.006 & \bf{55.56} & \bf{0.063} & 0.368 & \bf{0.692} & \bf{0.467} & \bf{0.600} & 0.412\\
FCF w0 c3 ( 90 x 1110 ) fused & \bf{0.568 ± 0.006} & 0.603 ± 0.005 & 51.11 & -0.016 & 0.368 & 0.615 & 0.412 & 0.571 & 0.389\\
FCF w5 c2 ( 135 x 1480 ) fused & 0.494 ± 0.005 & 0.526 ± 0.004 & 53.33 & 0.023 & 0.368 & 0.654 & 0.438 & 0.586 & 0.400\\
FCF w5 c3 ( 90 x 2220 ) fused & 0.483 ± 0.005 & 0.539 ± 0.007 & 53.33 & 0.023 & 0.368 & 0.654 & 0.438 & 0.586 & 0.400\\
\bottomrule
\end{tabular}
\end{table}

In the supervised setting (see Table~\ref{tab:sup-patient}), the best overall result was a mean AUC ROC of 0.715 without and a mean AUC ROC of 0.721 with decision fusion. The winning solution without fusion was \emph{RF w5 cms}, which uses the mean and standard deviation of 30 values (from 6 channels and 5 windows of each channel) for each audio feature, resulting in 740 features (370 means and 370 standard deviations). The winning solution with fusion was \emph{RF w5 c3}, which uses the mean and standard deviation of 5 values (from 5 windows) for each channel when calculating each audio feature, followed by the concatenation at the feature-level of these summary features from 3 channels side by side, and to obtain the final prediction for the subject separate predictions for right and left side are averaged out. In particular, twice as few features for concatenation (without the need for windows and, consequently, channel-level mean and standard deviation summary) in \emph{RF w0 c3} resulted in almost identical performance to the winner \emph{RF w5 c3} in decision fusion.

\begin{table}[!htb]
\scalefont{0.85}
\caption{\label{tab:sup-patient}Patient-based results in the supervised machine learning setting. Notes: bold denotes the maximum value of the column, and decision fusion term 'fused' corresponds to 'fused for Code'.}
\centering
\begin{tabular}[t]{lccccccccc}
\toprule
\bf{Model} & \bf{AUC ROC} & \bf{AUC PRC} & \bf{Acc} & \bf{Kappa} & \bf{Sens} & \bf{Spec} & \bf{Prec} & \bf{NPV} & \bf{F1} \\
\midrule
RF w0 cms ( 45 x 740 ) & 0.619 ± 0.008 & 0.486 ± 0.008 & 62.22 & 0.242 & 0.632 & 0.615 & 0.545 & 0.696 & 0.585\\
RF w3 cms ( 45 x 740 ) & 0.639 ± 0.009 & 0.522 ± 0.010 & 62.22 & 0.242 & 0.632 & 0.615 & 0.545 & 0.696 & 0.585\\
RF w5 cms ( 45 x 740 ) & \bf{0.715 ± 0.010} & \bf{0.609 ± 0.013} & \bf{68.89} & \bf{0.371} & \bf{0.684} & \bf{0.692} & \bf{0.619} & \bf{0.750} & \bf{0.650}\\
RF w0 c6 ( 45 x 2220 ) & 0.555 ± 0.016 & 0.440 ± 0.014 & 55.56 & 0.114 & 0.579 & 0.538 & 0.478 & 0.636 & 0.524\\
RF w5 c6 ( 45 x 4440 ) & 0.603 ± 0.016 & 0.480 ± 0.014 & 66.67 & 0.322 & 0.632 & 0.692 & 0.600 & 0.720 & 0.615\\
\midrule
RF w0 raw ( 270 x 373 ) fused & 0.621 ± 0.010 & 0.473 ± 0.010 & 64.44 & 0.281 & 0.632 & 0.654 & 0.571 & 0.708 & 0.600\\
RF w3 raw ( 810 x 373 ) fused & 0.655 ± 0.008 & 0.509 ± 0.010 & \bf{68.89} & \bf{0.371} & \bf{0.684} & \bf{0.692} & \bf{0.619} & \bf{0.750} & \bf{0.650}\\
RF w5 raw ( 1350 x 373 ) fused & 0.653 ± 0.008 & 0.508 ± 0.010 & 64.44 & 0.281 & 0.632 & 0.654 & 0.571 & 0.708 & 0.600\\
RF w5 wms ( 270 x 743 ) fused & 0.666 ± 0.009 & 0.523 ± 0.012 & \bf{68.89} & \bf{0.371} & \bf{0.684} & \bf{0.692} & \bf{0.619} & \bf{0.750} & \bf{0.650}\\
RF w0 c2 ( 135 x 741 ) fused & 0.597 ± 0.010 & 0.469 ± 0.010 & 60.00 & 0.192 & 0.579 & 0.615 & 0.524 & 0.667 & 0.550\\
RF w0 c3 ( 90 x 1111 ) fused & 0.715 ± 0.010 & \bf{0.562 ± 0.013} & \bf{68.89} & \bf{0.371} & \bf{0.684} & \bf{0.692} & \bf{0.619} & \bf{0.750} & \bf{0.650}\\
RF w5 c2 ( 135 x 1481 ) fused & 0.595 ± 0.010 & 0.469 ± 0.010 & 62.22 & 0.242 & 0.632 & 0.615 & 0.545 & 0.696 & 0.585\\
RF w5 c3 ( 90 x 2221 ) fused & \bf{0.721 ± 0.008} & \bf{0.562 ± 0.011} & \bf{68.89} & \bf{0.371} & \bf{0.684} & \bf{0.692} & \bf{0.619} & \bf{0.750} & \bf{0.650}\\
\bottomrule
\end{tabular}
\end{table}

\begin{figure}[!htb]
\centering
\includegraphics[width=.8\linewidth]{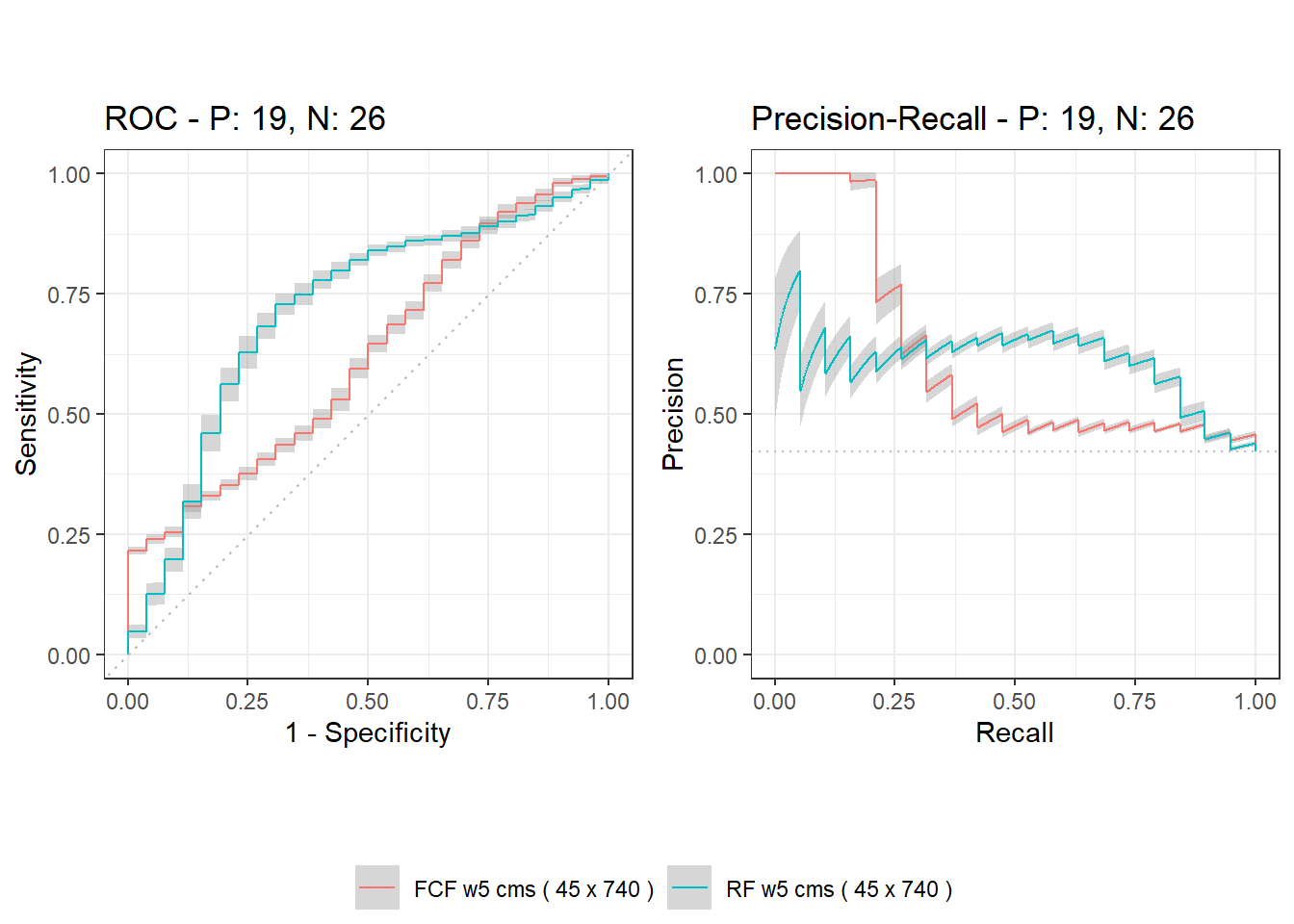}
\caption{\label{fig:patient} The best patient-based detection results summarized by the ROC and PRC curves for unsupervised (red) and supervised (blue) machine learning without decision fusion.}
\end{figure}

\begin{figure}[!htb]
\centering
\includegraphics[width=.8\linewidth]{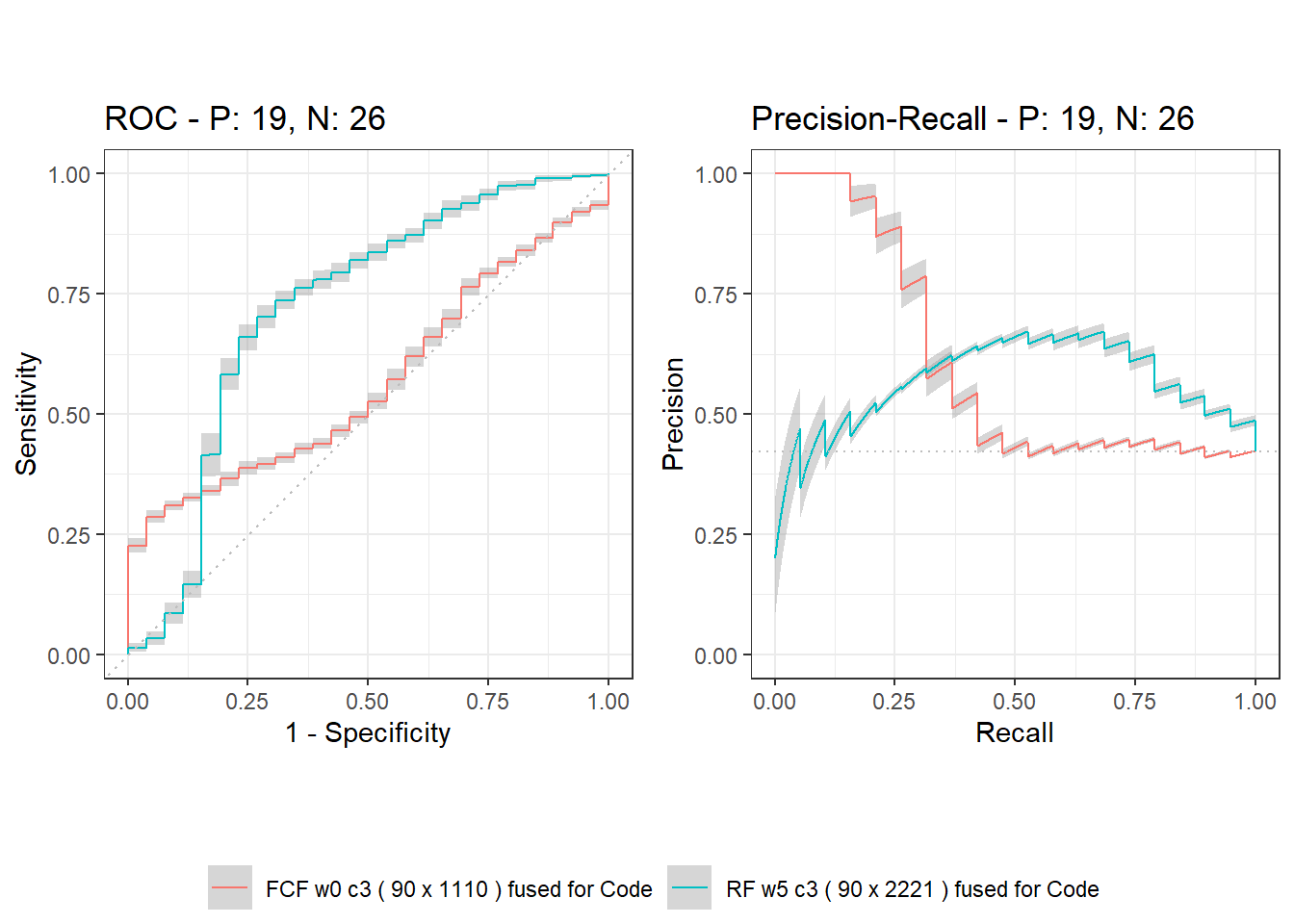}
\caption{\label{fig:patientFusion} The best patient-based detection results summarized by the ROC and PRC curves for unsupervised (red) and supervised (blue) machine learning with decision fusion.  }
\end{figure}

The performance curves (ROC and PRC) of the winning solutions for patient-based detection are shown in Figs. \ref{fig:patient}-\ref{fig:patientFusion}, where one can notice that supervised models clearly outperform unsupervised ones.

\subsection{Side-based detection results}

Side-based results constitute detection accuracy when two final decisions are independently obtained for a subject, for the left and right channels. Accuracy here is understood as a percentage of correctly predicted side-based decisions when the task is to distinguish between healthy and pathological cases.

In the unsupervised setting (see Table~\ref{tab:unsup-side}), the best overall result was a mean AUC ROC of 0.564 without and a mean AUC ROC of 0.555 with decision fusion. The winning solution without fusion was \emph{FCF w0 c3}, which uses side-by-side concatenation at the feature-level of 370 features from 3 channels, resulting in 1110 features. The winning solution with fusion was \emph{FCF w3 raw}, which obtains 9 predictions (3 windows in 3 channels on one side) for each subject and averages them for side-based decision fusion. Other fusion variants, both non-windowed \emph{FCF w0 raw} variant and variant \emph{FCF w5 raw} with more windows, were very close to the winning solution \emph{FCF w3 raw}.

\begin{table}[!htb]
\scalefont{0.85}
\caption{\label{tab:unsup-side}Side-based results in the unsupervised machine learning setting. Notes: bold denotes the maximum value of the column, and the decision fusion term 'fused' corresponds to 'fused for Side'.}
\centering
\begin{tabular}[t]{lccccccccc}
\toprule
\bf{Model} & \bf{AUC ROC} & \bf{AUC PRC} & \bf{Acc} & \bf{Kappa} & \bf{Sens} & \bf{Spec} & \bf{Prec} & \bf{NPV} & \bf{F1} \\
\midrule
FCF w0 c3 ( 90 x 1110 ) & \bf{0.564 ± 0.004} & \bf{0.543 ± 0.004} & \bf{55.56} & \bf{0.069} & 0.395 & \bf{0.673} & \bf{0.469} & \bf{0.603} & 0.429\\
FCF w5 c3 ( 90 x 2220 ) & 0.503 ± 0.004 & 0.475 ± 0.004 & 48.89 & 0.009 & \bf{0.605} & 0.404 & 0.426 & 0.583 & \bf{0.500}\\
\midrule
FCF w0 raw ( 270 x 370 ) fused & 0.545 ± 0.003 & 0.525 ± 0.003 & 47.78 & -0.010 & \bf{0.605} & 0.385 & 0.418 & 0.571 & \textbf{0.495}\\
FCF w3 raw ( 810 x 370 ) fused & \textbf{0.555 ± 0.002} & \textbf{0.555 ± 0.003} & \textbf{53.33} & \textbf{0.063} & 0.526 & \textbf{0.538} & \textbf{0.455} & \textbf{0.609} & 0.488\\
FCF w5 raw ( 1350 x 370 ) fused & 0.554 ± 0.002 & 0.554 ± 0.004 & 47.78 & -0.016 & 0.579 & 0.404 & 0.415 & 0.568 & 0.484\\
FCF w5 wms ( 270 x 740 ) fused & 0.505 ± 0.002 & 0.476 ± 0.003 & 52.22 & 0.038 & 0.500 & \textbf{0.538} & 0.442 & 0.596 & 0.469\\
\bottomrule
\end{tabular}
\end{table}

In the supervised setting (see Table~\ref{tab:sup-side}), the best overall result was a mean AUC ROC of 0.691 without and a mean AUC ROC of 0.645 with decision fusion. The winning solution without fusion was \emph{RF w5 c3}, which uses the mean and standard deviation of 5 values (from 5 windows) for each channel when calculating each audio feature, followed by the concatenation at the feature-level these summary features from 3 channels side by side, resulting in 2220 features (1110 means and 1110 standard deviations). In particular, half the number of features for concatenation (without the need for windows and consequently the channel-level summary by mean and standard deviation summary) in \emph{RF w0 c3} resulted in being very close to the winner. The winning solution with fusion was \emph{RF w5 wms}, which uses the mean and standard deviation of 5 values (from 5 windows) for each channel when calculating each audio feature, and the resulting 3 predictions for a side are averaged to obtain the final decision. Interestingly, other windowed solutions in supervised decision fusion – \emph{RF w3 raw} and \emph{RF w5 raw} – were relatively close to the winner \emph{RF w5 wms}.

\begin{table}[!htb]
\scalefont{0.85}
\caption{\label{tab:sup-side}Side-based results in the supervised machine learning setting. Notes: bold denotes the maximum value of the column, and the decision fusion term 'fused' corresponds to 'fused for Side'.}
\centering
\begin{tabular}[t]{lccccccccc}
\toprule
\bf{Model} & \bf{AUC ROC} & \bf{AUC PRC} & \bf{Acc} & \bf{Kappa} & \bf{Sens} & \bf{Spec} & \bf{Prec} & \bf{NPV} & \bf{F1} \\
\midrule
RF w0 c3 ( 90 x 1111 ) & 0.689 ± 0.010 & 0.546 ± 0.012 & 66.67 & 0.326 & 0.658 & 0.673 & 0.595 & 0.729 & 0.625\\
RF w5 c3 ( 90 x 2221 ) & \bf{0.691 ± 0.009} & \bf{0.553 ± 0.012} & \bf{71.11} & \bf{0.416} & \bf{0.711} & \bf{0.712} & \bf{0.643} & \bf{0.771} & \bf{0.675}\\
\midrule
RF w0 raw ( 270 x 373 ) fused & 0.607 ± 0.007 & 0.462 ± 0.006 & 54.44 & 0.089 & 0.553 & 0.538 & 0.467 & 0.622 & 0.506\\
RF w3 raw ( 810 x 373 ) fused & 0.641 ± 0.007 & 0.491 ± 0.007 & 58.89 & 0.172 & 0.579 & 0.596 & 0.512 & 0.660 & 0.543\\
RF w5 raw ( 1350 x 373 ) fused & 0.636 ± 0.008 & 0.498 ± 0.008 & 57.78 & 0.153 & 0.579 & 0.577 & 0.500 & 0.652 & 0.537\\
RF w5 wms ( 270 x 743 ) fused & \bf{0.645 ± 0.007} & \bf{0.512 ± 0.010} & \bf{64.44} & \bf{0.286} & \bf{0.658} & \bf{0.635} & \bf{0.568} & \bf{0.717} & \bf{0.610}\\
\bottomrule
\end{tabular}
\end{table}

\begin{figure}[!htb]
\centering
\includegraphics[width=.8\linewidth]{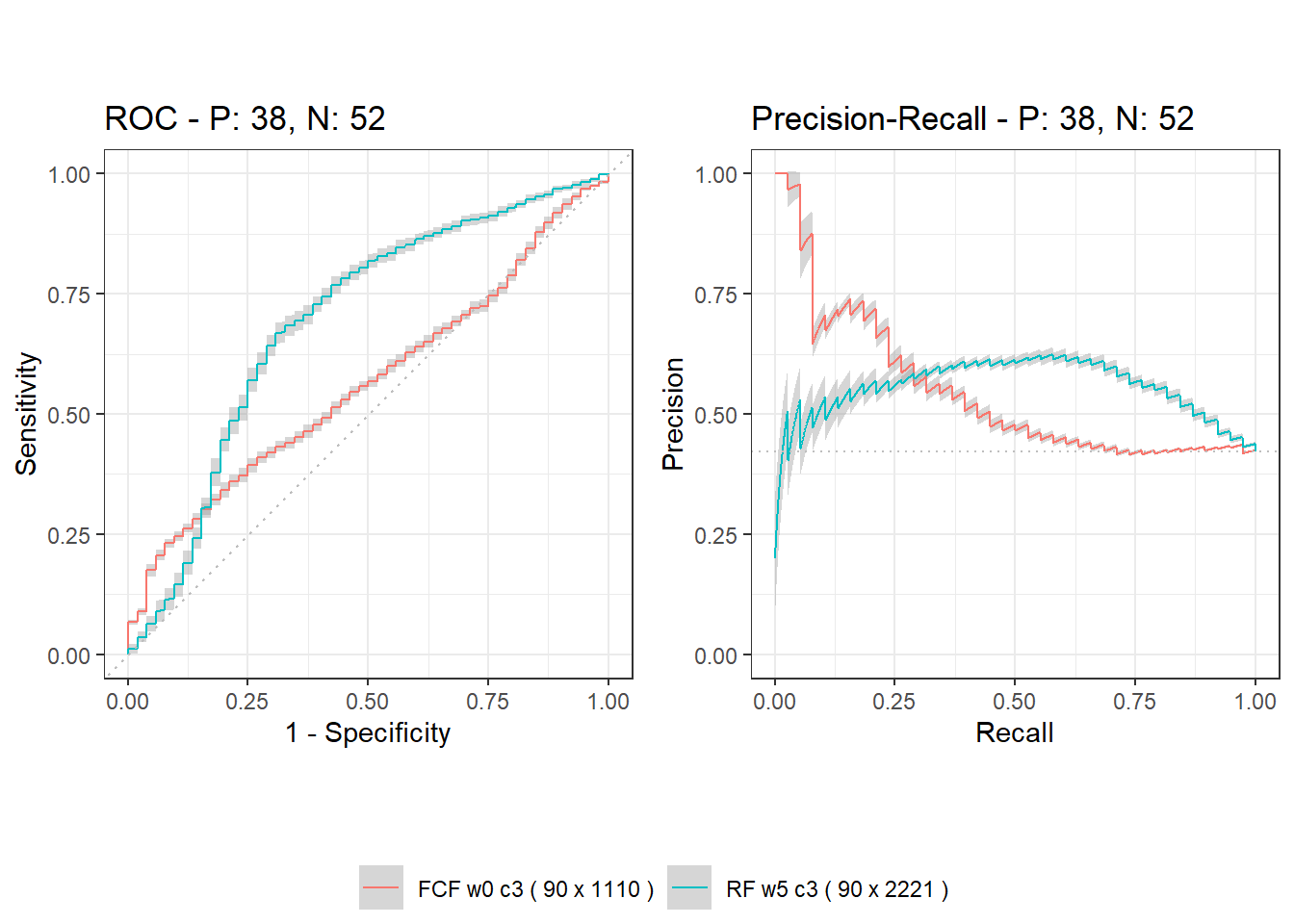}
\caption{\label{fig:side} The best side-based detection results summarized by the ROC and PRC curves for unsupervised (red) and supervised (blue) machine learning without decision fusion.}
\end{figure}

\begin{figure}[!htb]
\centering
\includegraphics[width=.8\linewidth]{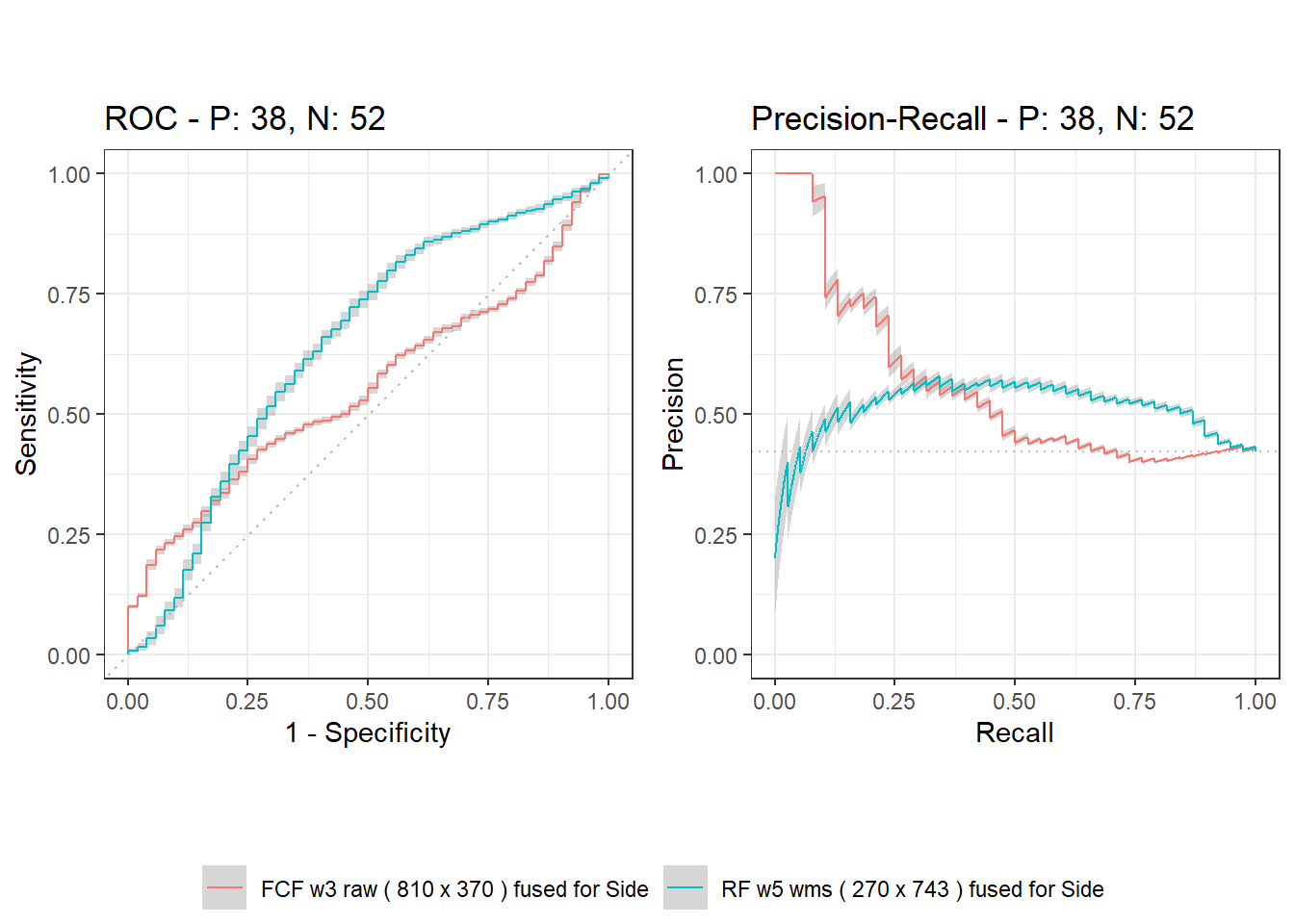}
\caption{\label{fig:sideFusion} The best side-based detection results summarized by the ROC and PRC curves for unsupervised (red) and supervised (blue) machine learning with decision fusion.}
\end{figure}

Performance curves (ROC and PRC) of the winner solutions for side-based detection are depicted in Figs. \ref{fig:side}-\ref{fig:sideFusion}, where one can notice that the supervised models clearly outperform the unsupervised ones.

\subsection{Level-based detection results}

Level-based results constitute detection accuracy when three decisions are obtained per subject. Accuracy here is understood as a percentage of correctly predicted level-based decisions when the task is to distinguish between healthy and pathological cases.

In the unsupervised setting (see Table~\ref{tab:unsup-level}), the best overall result was a mean AUC ROC of 0.530 without and a mean AUC ROC of 0.532 with decision fusion. The winning solution without fusion was \emph{FCF w0 c2}, which uses side-by-side concatenation at the feature-level for 2 channels of the same level, resulting in 740 features. The winning solution with fusion was \emph{FCF w5 raw}, which obtains 10 predictions (5 windows in 2 channels on one level) and averages them for level-based decision fusion. Most of the other solutions in the unsupervised decision fusion – \emph{FCF w0 raw} and \emph{FCF w3 raw} – were rather close to the winner \emph{FCF w5 raw}.

\begin{table}[!htb]
\scalefont{0.85}
\caption{\label{tab:unsup-level}Level-based results in unsupervised machine learning setting. Notes: bold denotes the maximum value of the column, and the decision fusion term 'fused' corresponds to 'fused for Level'.}
\centering
\begin{tabular}[t]{lccccccccc}
\toprule
\bf{Model} & \bf{AUC ROC} & \bf{AUC PRC} & \bf{Acc} & \bf{Kappa} & \bf{Sens} & \bf{Spec} & \bf{Prec} & \bf{NPV} & \bf{F1} \\
\midrule
FCF w0 c2 ( 135 x 740 ) & \textbf{0.530 ± 0.004} & \textbf{0.488 ± 0.004} & \textbf{52.59} & \textbf{0.024} & \textbf{0.421} & \textbf{0.603} & \textbf{0.436} & \textbf{0.587} & \textbf{0.429}\\
FCF w5 c2 ( 135 x 1480 ) & 0.495 ± 0.003 & 0.463 ± 0.003 & 51.11 & -0.012 & 0.386 & \textbf{0.603} & 0.415 & 0.573 & 0.400\\
\midrule
FCF w0 raw ( 270 x 370 ) fused & 0.527 ± 0.003 & 0.512 ± 0.002 & 49.63 & 0.019 & \textbf{0.596} & 0.423 & 0.430 & 0.589 & \textbf{0.500}\\
FCF w3 raw ( 810 x 370 ) fused & 0.530 ± 0.002 & 0.520 ± 0.002 & \textbf{51.85} & \textbf{0.038} & 0.526 & \textbf{0.513} & \textbf{0.441} & \textbf{0.597} & 0.480\\
FCF w5 raw ( 1350 x 370 ) fused & \textbf{0.532 ± 0.002} & \textbf{0.522 ± 0.002} & 48.15 & -0.015 & 0.561 & 0.423 & 0.416 & 0.569 & 0.478\\
FCF w5 wms ( 270 x 740 ) fused & 0.486 ± 0.002 & 0.462 ± 0.002 & 49.63 & -0.009 & 0.491 & 0.500 & 0.418 & 0.574 & 0.452\\
\bottomrule
\end{tabular}
\end{table}

In the supervised setting (see Table~\ref{tab:sup-level}), the best overall result was a mean AUC ROC of 0.581 without and a mean AUC ROC of 0.648 with decision fusion. The winning solution without fusion was \emph{RF w5 c2}, which uses the mean and standard deviation of 5 values (from 5 windows) for each channel when calculating each audio feature, followed by side-by-side concatenation at the feature-level of 2 channels. The winning solution with fusion was \emph{RF w5 wms}.

\begin{table}[!htb]
\scalefont{0.85}
\caption{\label{tab:sup-level}Level-based results in supervised machine learning setting. Notes: bold denotes the maximum value of the column, and the decision fusion term 'fused' corresponds to 'fused for Level'.}
\centering
\begin{tabular}[t]{lccccccccc}
\toprule
\bf{Model} & \bf{AUC ROC} & \bf{AUC PRC} & \bf{Acc} & \bf{Kappa} & \bf{Sens} & \bf{Spec} & \bf{Prec} & \bf{NPV} & \bf{F1} \\
\midrule
RF w0 c2 ( 135 x 741 ) & 0.579 ± 0.010 & \textbf{0.458 ± 0.009} & \textbf{61.48} & \textbf{0.225} & \textbf{0.614} & \textbf{0.615} & \textbf{0.538} & \textbf{0.686} & \textbf{0.574}\\
RF w5 c2 ( 135 x 1481 ) & \textbf{0.581 ± 0.009} & 0.454 ± 0.007 & 60.00 & 0.195 & 0.596 & 0.603 & 0.523 & 0.671 & 0.557\\
\midrule
RF w0 raw ( 270 x 373 ) fused & 0.606 ± 0.007 & 0.473 ± 0.007 & 58.52 & 0.169 & 0.596 & 0.577 & 0.507 & 0.662 & 0.548\\
RF w3 raw ( 810 x 373 ) fused & 0.627 ± 0.007 & 0.489 ± 0.007 & 62.22 & 0.238 & 0.614 & \bf{0.628} & 0.547 & 0.690 & 0.579\\
RF w5 raw ( 1350 x 373 ) fused & 0.627 ± 0.007 & 0.493 ± 0.008 & 58.52 & 0.166 & 0.579 & 0.590 & 0.508 & 0.657 & 0.541\\
RF w5 wms ( 270 x 743 ) fused & \textbf{0.648 ± 0.007} & \textbf{0.507 ± 0.007} & \textbf{63.70} & \textbf{0.272} & \textbf{0.649} & \textbf{0.628} & \textbf{0.561} & \textbf{0.710} & \textbf{0.602}\\
\bottomrule
\end{tabular}
\end{table}

\begin{figure}[!htb]
\centering
\includegraphics[width=.8\linewidth]{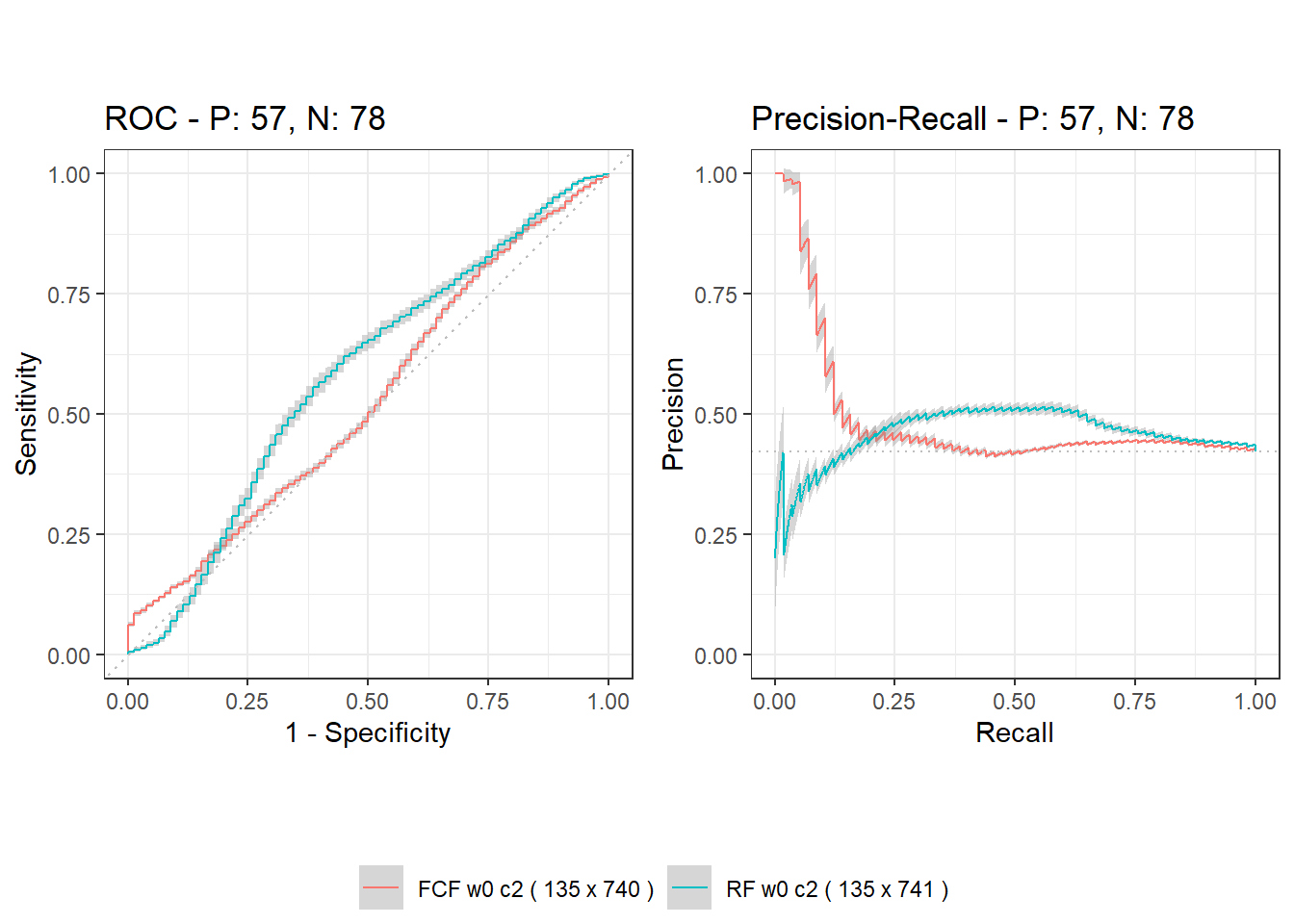}
\caption{\label{fig:level} The best level-based detection results summarized by the ROC and PRC curves for unsupervised (red) and supervised (blue) machine learning without decision fusion.}
\end{figure}

\begin{figure}[!htb]
\centering
\includegraphics[width=.8\linewidth]{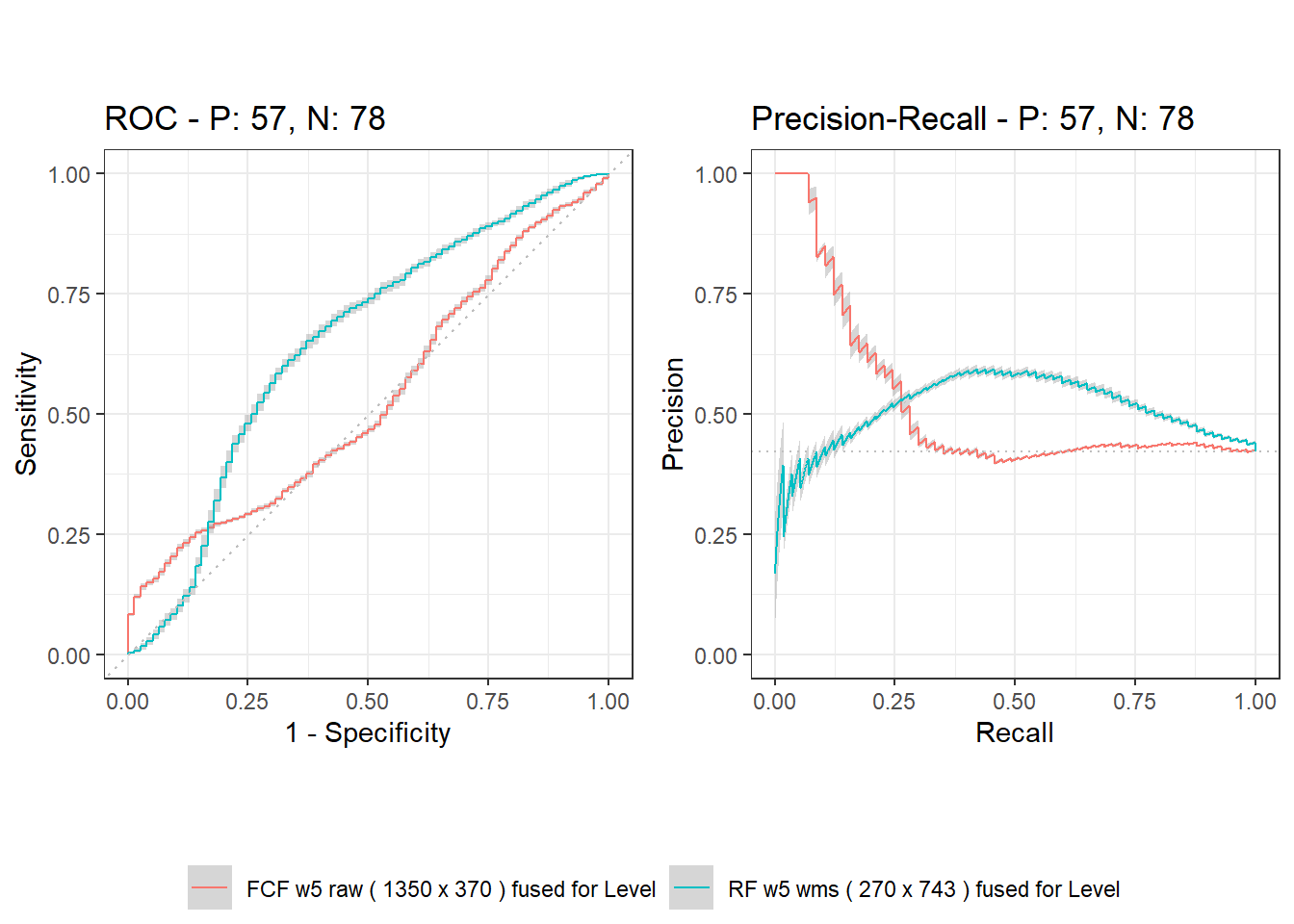}
\caption{\label{fig:levelFusion} The best level-based detection results summarized by the ROC and PRC curves for unsupervised (red) and supervised (blue) machine learning with decision fusion.  }
\end{figure}

The performance curves (ROC and PRC) of the winner solutions for level-based detection are depicted in Figs. \ref{fig:level}-\ref{fig:levelFusion}, where one can notice that the supervised models clearly outperform the unsupervised ones.

\subsection{Channel-based / window-based detection results}

Channel-based results constitute detection accuracy when 6 decisions (for a dataset of 270 rows) are obtained per subject. Window-based results constitute detection accuracy when 18 (for a dataset of 810 rows) or 30 (for a dataset of 1350 rows) decisions are obtained per subject.

In the unsupervised setting (see Table \ref{tab:unsup-channel}), the best overall result was a mean AUC ROC of 0.536 both without and with decision fusion. The channel-based solution without fusion that won was \emph{FCF w0 raw}, which contains 370 basic features and obtains a prediction for each channel. The window-based solution without fusion that won -- \emph{FCF w3 raw} -- performed rather similarly to the channel-based counterpart \emph{FCF w0 raw}. The channel-based solution with fusion that won was \emph{FCF w3 raw}, which obtains 18 predictions per subject and then averages three window-level predictions for each of the six channels. Decision fusion based on the average of predictions for five windows -- \emph{FCF w5 raw} -- was also very close to the winner.

\begin{table}[!htb]
\scalefont{0.85}
\caption{\label{tab:unsup-channel}Channel-based and window-based results in the supervised machine learning setting. Notes: bold denotes the maximum value of the column, and the decision fusion term 'fused' corresponds to 'fused for Channel'.}
\centering
\begin{tabular}[t]{lccccccccc}
\toprule
\bf{Model} & \bf{AUC ROC} & \bf{AUC PRC} & \bf{Acc} & \bf{Kappa} & \bf{Sens} & \bf{Spec} & \bf{Prec} & \bf{NPV} & \bf{F1} \\
\midrule
FCF w0 raw ( 270 x 370 ) & \bf{0.536 ± 0.002} & \textbf{0.469 ± 0.002} & 52.59 & 0.059 & \textbf{0.561} & 0.500 & 0.451 & \textbf{0.609} & \textbf{0.500}\\
FCF w5 wms ( 270 x 740 ) & 0.505 ± 0.001 & 0.431 ± 0.001 & 50.37 & -0.008 & 0.447 & 0.545 & 0.418 & 0.574 & 0.432\\
FCF w3 raw ( 810 x 370 ) & 0.535 ± 0.002 & 0.467 ± 0.001 & \textbf{53.95} & \textbf{0.062} & 0.480 & \textbf{0.583} & \textbf{0.457} & 0.605 & 0.468\\
FCF w5 raw ( 1350 x 370 ) & 0.530 ± 0.001 & 0.461 ± 0.001 & 51.11 & 0.024 & 0.519 & 0.505 & 0.434 & 0.590 & 0.473\\
\midrule
FCF w3 raw ( 810 x 370 ) fused & \textbf{0.536 ± 0.002} & 0.479 ± 0.001 & \textbf{55.56} & \textbf{0.098} & 0.509 & \textbf{0.590} & \textbf{0.475} & \textbf{0.622} & 0.492\\
FCF w5 raw ( 1350 x 370 ) fused & 0.534 ± 0.001 & \textbf{0.480 ± 0.002} & 53.33 & 0.078 & \textbf{0.588} & 0.494 & 0.459 & 0.621 & \textbf{0.515}\\
\bottomrule
\end{tabular}
\end{table}

In the supervised setting (see Table \ref{tab:sup-channel}), the best result was a mean AUC ROC of 0.618 without and a mean AUC ROC of 0.607 with decision fusion. The channel-based solution that won without fusion was \emph{RF w5 wms}, which uses the mean and standard deviation of 5 values for each channel. The window-based solution without fusion \emph{RF w3 raw} was slightly worse than the channel-based winner.  The winning solution with fusion was \emph{RF w5 raw}, which averages five predictions for each channel. Although using fewer windows in decision fusion -- \emph{RF w3 raw} -- when three predicions are averaged, resulted in a very similar performance.

\begin{table}[!htb]
\scalefont{0.85}
\caption{\label{tab:sup-channel}Channel-based and window-based results in the supervised machine learning setting. Notes: bold denotes the maximum value of the column, and the decision fusion term 'fused' corresponds to 'fused for Channel'.}
\centering
\begin{tabular}[t]{lccccccccc}
\toprule
\bf{Model} & \bf{AUC ROC} & \bf{AUC PRC} & \bf{Acc} & \bf{Kappa} & \bf{Sens} & \bf{Spec} & \bf{Prec} & \bf{NPV} & \bf{F1} \\
\midrule
RF w0 raw ( 270 x 373 ) & 0.584 ± 0.007 & 0.463 ± 0.007 & 56.67 & 0.131 & 0.570 & 0.564 & 0.489 & 0.642 & 0.526\\
RF w5 wms ( 270 x 743 ) & \textbf{0.618 ± 0.006} & \textbf{0.492 ± 0.007} & \textbf{58.89} & \textbf{0.176} & \textbf{0.596} & 0.583 & \textbf{0.511} & \textbf{0.664} & \textbf{0.551}\\
RF w3 raw ( 810 x 373 ) & 0.598 ± 0.005 & 0.469 ± 0.005 & 57.90 & 0.152 & 0.567 & \textbf{0.588} & 0.501 & 0.650 & 0.532\\
RF w5 raw ( 1350 x 373 ) & 0.595 ± 0.005 & 0.474 ± 0.006 & 57.41 & 0.146 & 0.577 & 0.572 & 0.496 & 0.649 & 0.534\\
\midrule
RF w3 raw ( 810 x 373 ) fused & \textbf{0.607 ± 0.006} & 0.476 ± 0.006 & 57.04 & 0.138 & 0.570 & 0.571 & 0.492 & 0.645 & 0.528\\
\addlinespace
RF w5 raw ( 1350 x 373 ) fused & 0.606 ± 0.006 & \textbf{0.485 ± 0.007} & \textbf{60.00} & \textbf{0.195} & \textbf{0.596} & \textbf{0.603} & \textbf{0.523} & \textbf{0.671} & \textbf{0.557}\\
\bottomrule
\end{tabular}
\end{table}

Due to the rather low mean AUC values for both the channel-based and window-based detection performance curves (ROC and PRC), the figures are not provided, but by comparing Table \ref{tab:unsup-channel} with Table \ref{tab:sup-channel} one can notice that the supervised models clearly outperform the unsupervised ones.

\section{Conclusions}

The prevalence of auscultation and digital stethoscopes in light of imprecision in physician evaluation makes this diagnostic tool an interesting research area with respect to the applicability of machine learning. Existing research shows that artificial intelligence has the potential to outperform medical specialists in terms of pathology detection accuracy.

In the experiments performed here, supervised machine learning methods consistently outperformed unsupervised ones, which is not surprising for small datasets. Random forest achieved a mean AUC ROC of 0.691 (accuracy 71.11\%, Kappa 0.416, and F1-score 0.771) in side-based detection and a mean AUC ROC of 0.721 (accuracy 68.89\%, Kappa 0.371, and F1-score 0.650) in patient-based detection. Splitting 15 s recording into five overlapping windows of 5 s in length, aggregating audio features from windows and/or channels with mean and standard deviation, and decision fusion by averaging several predictions for a subject were simple but useful techniques for improving detection success.

Our pilot research shows that supervised machine learning using the standard audio feature extractor can be successfully applied to differentiate between pathological and normal lung sounds. The further research direction could be to expand the corpus with more patients, extend the scope of diagnosis with heart pathologies, experiment with the latest sound corpus \emph{HF\_Lung\_V2} \cite{Hsu_2022}, and apply deep learning to spectrogram-based image representations of sounds with the aim of improving detection success.

\section*{Acknowledgements}

This work was supported by a student summer research grant provided to Lukas Drukteinis and Evaldas Vaičiukynas by the Research Council of Lithuania (agreement No. P-SV-22-241). Haroldas Razvadauskas is grateful to physician Jurgita Razvadauskienė for advice and support. We also would like to thank the heads of departments and their staff of physicians and nurses for allowing and assisting to enroll subjects for auscultation, more notably: Laima Jankauskienė (head of Department of Cardiology), Irena Šidiškienė (head of Department of Internal Medicine), and Inga Beinaravičienė (head of Department of Internal Medicine).

\bibliographystyle{abbrvnat}


\end{document}